# Unsupervised Lexical Acquisition of Relative Spatial Concepts Using Spoken User Utterances

Rikunari Sagara[1], Ryo Taguchi[1], Akira Taniguchi[2], Tadahiro Taniguchi[2], Koosuke Hattori[3], Masahiro Hoguro[3] and Taizo Umezaki[3]

[1]*Taguchi Laboratory, Department of Computer Science, Nagoya Institute of Technology, Aichi, Japan*

[2]*Emergent Systems Laboratory, College of Information Science and Engineering, Ritsumeikan University, Shiga, Japan*

[3]*College of Engineering, Chubu University, Aichi, Japan*

**Correspondence:**
Rikunari Sagara
r.sagara.628@nitech.jp

# Unsupervised Lexical Acquisition of Relative Spatial Concepts Using Spoken User Utterances

This paper proposes methods for unsupervised lexical acquisition for relative spatial concepts using spoken user utterances. A robot with a flexible spoken dialog system must be able to acquire linguistic representation and its meaning specific to an environment through interactions with humans as children do. Specifically, relative spatial concepts (e.g., front and right) are widely used in our daily lives, however, it is not obvious which object is a reference object when a robot learns relative spatial concepts. Therefore, we propose methods by which a robot without prior knowledge of words can learn relative spatial concepts. The methods are formulated using a probabilistic model to estimate the proper reference objects and distributions representing concepts simultaneously. The experimental results show that relative spatial concepts and a phoneme sequence representing each concept can be learned under the condition that the robot does not know which located object is the reference object. Additionally, we show that two processes in the proposed method improve the estimation accuracy of the concepts: generating candidate word sequences by class n-gram and selecting word sequences using location information. Furthermore, we show that clues to reference objects improve accuracy even though the number of candidate reference objects increases.

Keywords: ambiguous speech recognition; Bayesian nonparametrics; lexical acquisition; spatial concept acquisition; relative concept acquisition

## 1  Introduction

Service robots that support human activities in human living environments and offices are required to have a flexible spoken dialog capability that can handle out-of-vocabulary (OOV) words. In a human living environment or an office, objects, places, tasks, and linguistic representations vary from environment to environment. It is therefore difficult for developers of robots to implement knowledge in advance for all the things that might be encountered in a specific environment. However, children can learn the meaning of words flexibly [1]–[3]. If a robot can learn the meaning of words from natural utterances as with children, the problem mentioned above can be solved [4]. To learn from natural utterances, word boundaries must be estimated simultaneously. In this study, this task is called lexical acquisition by robots. Studies have been conducted on this topic [5]. For example, Taniguchi et al. proposed a spatial concept acquisition method (SpCoA++), which is a model for integrating locations and spoken user utterances [6]. They extended the model by involving and proposing SpCoSLAM [7][8]. The acquired lexicon and spatial concept can also be used for navigation tasks [9]. Their methods learn areas 'in front of the table' and 'in front of the TV' as separate concepts because they learn spatial concepts as absolute areas on the map. If the concepts can be learned as a generalized relative concept 'in front of ...' a robot can understand a new environment more flexibly. The relative concepts depend on the reference points. For example, the trainer utters unknown words, 'baba' and 'kiki,' which represent a relative concept and a reference point respectively, in a situation illustrated in Fig. 1. In this case, 'baba' means 'front' if TV is the reference point. On the other hand, 'baba' means 'left' if the chair is the reference point. Therefore, it must be known which of these, the TV or the chair, is selected as the reference point to learn the meaning of 'baba.' However, a reference point might be omitted in a human utterance such as 'The store is on the left side.' In addition, when acquiring language, even if the trainer utters the word of the object selected as the reference point such as 'kiki,' the word might not be understood. Therefore, the learning of relative concepts must be performed flexibly by estimating a reference point selected by a speaker in each scene.

In this study, we propose novel methods and examined how robots learn relative spatial concepts, which are important for human-robot interactions. Using the proposed methods, a robot without prior knowledge of words can learn relative spatial concepts by estimating a suitable *reference object*, defined as an object used as a reference point. First, we propose a relative spatial concept acquisition method (ReSCAM). Relative Spatial concepts and words representing them are learned from utterances without words representing reference objects. Therefore, using ReSCAM, reference objects are estimated so that consistency in learning results is ensured. Next, we propose a relative spatial concept acquisition method using clues to reference objects (ReSCAM+O). Concepts are learned from utterances with words representing reference objects. The robot needs to learn which object is represented by the word in the utterance because the robot has no prior knowledge of words. Words representing objects and relative spatial concepts can be learned simultaneously using ReSCAM+O. As we described, because concepts are learned with no prior knowledge of words, the proposed methods are not dependent on a specific language.

The main contributions of this paper are as follows:

- We propose novel methods: ReSCAM and ReSCAM+O, and show that these methods can segment words with high accuracy and learn relative spatial concepts.
- We show that the estimation accuracy of relative spatial concepts can be improved by class n-grams and mutual information.
- We demonstrate that the estimation accuracy can be improved by clues to reference objects: object recognition results and words representing objects.

The remainder of this paper is organized as follows. In Section 2, we discuss previous studies relevant to our study. In Section 3 and 4, we present our proposed method ReSCAM and ReSCAM+O. In Section 5 and 6, we present the experimental results obtained with ReSCAM and ReSCAM+O. In Section 7, we present conclusions drawn from the results of the study.

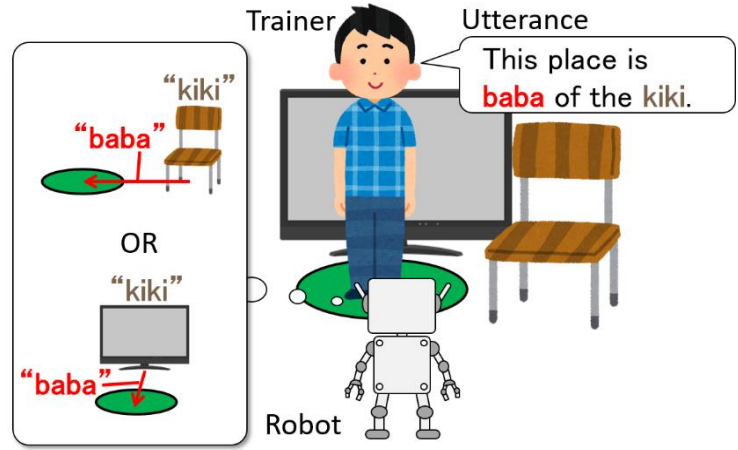

Figure 1. An interaction scene for learning relative spatial concepts. We assume that words 'baba' and 'kiki' are unknown words.

## 2 Related work

Our work integrates two tasks: lexical acquisition and learning of concepts dependent on reference points. We discuss related work on each task.

### 2.1 Lexical acquisition of absolute concepts

Frank et al. proposed a Bayesian model for cross-situational word learning [10]. This model clarifies the understanding of word learning. Their paper mentions that a Bayesian model can be easily extended for joint learning with other domains. For this reason, we also used a Bayesian model for learning concepts. Heath et al. proposed a method for mobile robots to learn lexical knowledge through robot-to-robot communication [11]. They showed that their method could resolve referential uncertainty for the dimensions of space and time. However, their method was not intended for lexical acquisition using human-to-robot speech interactions. Hatori et al. devised a comprehensive system that can accommodate unconstrained spoken language and effectively resolve ambiguity in spoken instruction [12]. In their study, the meaning of words, 'bottom-right,' is learned. However, these spatial concepts are not relative concepts because they do not depend on the reference points. Štepánová et al. proposed a computational method for mapping language to vision using a real-world robotic scenario [13]. This experimental result showed that the method could find mapping between language and vision robustly. However, words were learned using a pre-trained language model. Taniguchi et al. proposed a Bayesian model that can form multiple categories based on sensory channels and can associate words with sensory channels [14]. However, in addition to the methods mentioned above, this method cannot learn phoneme sequences of unknown words from spoken utterances because it presupposes the input of words.

On the other hand, Heymann et al. proposed a method that alternately updates phoneme recognition results and a language model using the unsupervised word segmentation method [15]. They showed that the accuracy of word segmentation could be improved using this method. However, they used only spoken user utterances rather than other information with high co-occurrence (e.g., place or object). We reason that the accuracy of lexical acquisition can be improved by adding other information as the input. Synnaeve et al. proposed word segmentation models that take the non-linguistic context into account [16]. Experimental results showed that their model produced better segmentation scores than its context-oblivious counterpart. However, labels used as context annotations are needed to apply this method. When a robot learns concepts from sensory information, it may fail to estimate labels. In our model, these labels are estimated from sensory information and word sequences to perform mutual complementation of ambiguity. Araki et al. proposed a learning method for object concepts and word meanings using multimodal information and spoken sentences [17]. Similarly, Nakamura et al. proposed a mutual learning method based on integrating the learning of object concepts with a language model [18]. In these methods, spoken sentences were segmented using an unsupervised morphological analyzer based on a nested Pitman–Yor language model (NPYLM) [19]. However, using NPYLM, word segmentation results include many incorrect boundaries if the recognized phoneme sequences have errors. To solve this problem, Taniguchi et al. proposed the SpCoA++ [6] adopting Neubig's unsupervised word segmentation method [20] which uses speech recognition lattices. We also used a segmentation method to solve the problem.

These studies focus on learning lexicons regarding absolute concepts (e.g., 'elevator hall'), which do not represent relative properties from reference points (e.g., 'in front of…'). We propose learning methods of relative concepts that are more difficult to learn than absolute concepts because we must estimate reference points.

## 2.2  *Learning of concepts dependent on reference points*

Sugiura et al. proposed a method for learning spatial moving concepts dependent on reference points by estimating coordinate systems and reference points using an expectation–maximization (EM) algorithm [21]. Similarly, Gu et al. proposed a method based on an EM algorithm to estimate reference points and learn relative spatial concepts [22]. In addition, Barrett et al. presented a unified framework supporting learning, generation, and comprehension of robotic driving [23]. Performances of these tasks were evaluated highly by human judges. However, in these studies, an utterance used for learning consisted of only one word. Aly et al. suggested a probabilistic framework for learning words representing spatial concepts (spatial prepositions) and object categories based on visual cues representing spatial layouts and geometric characteristics of objects in a tabletop scene [24]. However, segmented word sequences are used for teaching. Therefore, concepts cannot be learned from spoken utterances, which is a more natural approach to teaching to a robot. In our study, we use spoken utterances for input so that concepts are learned in closer settings to real teaching.

Spranger et al. put forth a learning method involving relative spatial concepts, similar to our study [25],[26]. They showed that relative spatial concepts in different coordinate systems can be learned by this method. However, they did not consider situations with several candidate reference objects. In that situation, our methods can estimate reference objects and learn concepts.

## 3  **Relative spatial concept acquisition method (ReSCAM)**

The proposed method ReSCAM enables robots to learn relative spatial concepts using spoken user utterances. This method is based on a probabilistic model.

### *3.1   Task setting*

We illustrate an interaction scene for relative spatial concept learning in Fig. 2. As all the experiments in this paper were performed in Japanese language, we show English translations as well as Japanese phoneme sequences in this paper. A trainer and a robot are in the scene, as well as candidates for reference objects. The trainer selects one object as a reference object from among the candidates and teaches the robot what his (the trainer's) position is relative to the reference object by uttering words. We assume that the trainer always utters from the robot's perspective[1].

The robot has an acoustic model and a language model of Japanese syllables as its initial knowledge and can recognize an utterance as a phoneme sequence. However, the robot cannot know the boundaries of words because the robot has no pre-existing lexicon. In addition, the robot cannot know which object is selected as the reference object. Such teaching is iterated several times by changing the positions of the trainer, the robot, and the objects. From the phoneme sequences and the relative locations of the trainer, the robot learns relative spatial concepts and words representing them while estimating the reference object selected by the trainer in each scene.

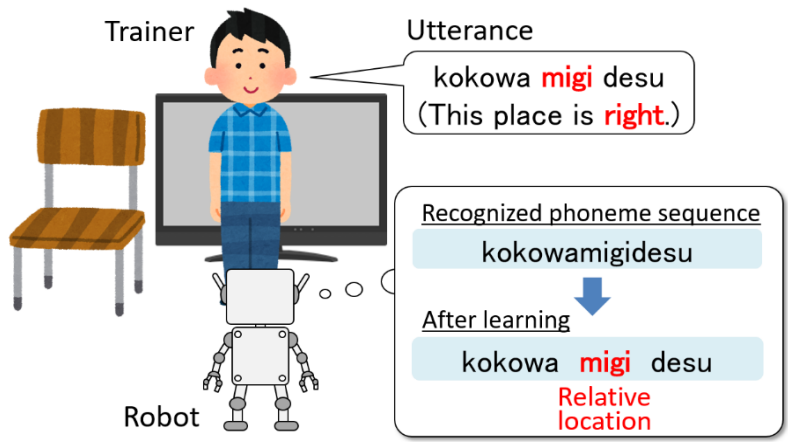

Figure 2. An interaction scene for lexical acquisition of relative spatial concepts.

### *3.2   Probabilistic generative model*

The process for representing relative locations by utterances is formulated by a generative model. A graphical model of the proposed method is shown in Fig. 3. A list of distributions is given in Table 1. A list of variables is given in Table 2. This model represents the co-occurrence of the spoken user utterance $y_n$ and the relative location of the trainer. This relative location is calculated from the absolute location of the trainer, the robot and the reference object.

---

[1] The relative location depends on the reference object and the coordinate system selected. If the trainer uses his own perspective, the utterance in Figure 2 can be 'left (of the chair)' instead of 'right.' This paper does not address estimation of the coordinate system to focus on estimation of reference objects. We consider that the robot can estimate coordinate systems by adopting methods proposed in Sugiura's study [21] and Gu's study [22].

The generative model of $x_n$ and $y_n$ is described below. The location of the trainer $x_n$ is defined as follows:

$$x_n = x^O_{n\pi_n} + R(f^O_{n\pi_n})x'_{n\pi_n} \quad (1)$$

where $R(\theta)$ denotes the rotation matrix of the angle $\theta$. In this study, we address projective concepts that represent directions in relative spatial concepts. Therefore, in this model, the relative location $x'_{nj}$ is determined by the distance $l_{nj}$ and angle $\theta_{nj}$. First, $l_{nj}$ denotes the distance between a reference object and trainer. We consider that the distance is fixed to some extent when the trainer teaches the concepts. Therefore, the distance is generated using a normal distribution called distance distribution. Additionally, $\theta_{nj}$ denotes the angle between a line that passes through a reference object and the trainer, and a line that passes through a reference object and a robot. The angle $\theta_{nj}$ is generated using a von Mises distribution that can represent the angles or directions. This distribution is called directional distribution. The relative location $x'_{nj}$ is generated as follows:

$$x'_{nj} \sim \begin{cases} N(l_{nj}|\mu, \lambda^{-1})vM(\theta_{nj}|v_{C^R_n}, \kappa_{C^R_n}) & (j = \pi_n) \\ U(l_{nj}|0, e)U(\theta_{nj}|0, 2\pi) & \text{(otherwise)} \end{cases} \quad (2)$$

This formulation means that the trainer's relative location $x'_{n\pi_n}$ is stochastically generated by the distance distribution $N(\cdot)$ and the directional distribution $vM(\cdot)$, which are conditioned by the relative spatial concept $C^R_n$ and parameters $\mu, \lambda^{-1}$ shared among relative spatial concepts.

The parameters of equation (2) are generated as follows:

$$\mu \sim N(\mu_0, \lambda_0) \quad (3)$$
$$\lambda \sim \text{Gam}(a_0, b_0) \quad (4)$$
$$v_s \sim vM(v_0, \kappa_0) \quad (5)$$
$$\kappa_s \sim \log N(m_0, \sigma_0) \quad (6)$$

The parameter $\kappa_s$ is generated from a log-normal distribution as in the case of Qin's model [27], because a conjugate prior of the parameter does not exist. Indices of reference objects $\pi_n$ and relative spatial concepts $C^R_n$ are generated as follows:

$$\pi_n \sim \text{Mult}(\gamma^\pi) \quad (7)$$
$$C^R_n \sim \text{CRP}(\alpha^R) \quad (8)$$

Using Chinese restaurant process (CRP) [28], the number of relative spatial concepts can be estimated while learning them. A word sequence $w_n$ is generated using approximation by unigram rescaling [29], as follows:

$$w_n \sim p(w_n|\phi^R, \psi, C^R_n, z^R_n, \mathcal{L})$$
$$\stackrel{UR}{\approx} p(w_n|\mathcal{L}) \prod_i \frac{p(w_{ni}|\phi^R, \psi, C^R_n, z^R_n)}{p(w_{ni})} \quad (9)$$

where $\boldsymbol{\phi}^R = \{\boldsymbol{\phi}_1^R, \ldots, \boldsymbol{\phi}_S^R\}$, and $\stackrel{UR}{\approx}$ denotes the approximation using unigram rescaling. $p(w_{ni}|\boldsymbol{\phi}^R, \boldsymbol{\psi}, C_n^R, z_n^R)$ denotes the prior probability of $w_{ni}$, the $i$th word of the word sequence $\boldsymbol{w}_n$. It is calculated as follows:

$$p(w_{ni}|\boldsymbol{\phi}^R, \boldsymbol{\psi}, C_n^R, z_n^R) = \begin{cases} \text{Mult}\left(w_{ni}|\boldsymbol{\phi}_{C_n^R}^R\right) & (i = z_n^R) \\ \text{Mult}(w_{ni}|\boldsymbol{\psi}) & (\text{otherwise}) \end{cases} \quad (10)$$

We define a *location word* as a word representing a relative spatial concept and a *concept-independent word* as a word that does not represent any concept we address in this study. The parameters of word distributions $\boldsymbol{\phi}_s^R, \boldsymbol{\psi}$ and indices of location words $z_n^R$ are generated as follows:

$$\boldsymbol{\phi}_s^R \sim \text{Dir}(\boldsymbol{\beta}^R) \quad (11)$$

$$\boldsymbol{\psi} \sim \text{Dir}(\boldsymbol{\beta}^\psi) \quad (12)$$

$$z_n^R \sim \text{Mult}(\boldsymbol{\gamma}^z) \quad (13)$$

An utterance $y_n$ is generated using a word sequence $\boldsymbol{w}_n$ as follows:

$$y_n \sim p(y_n|\boldsymbol{w}_n, \mathcal{A}) \quad (14)$$

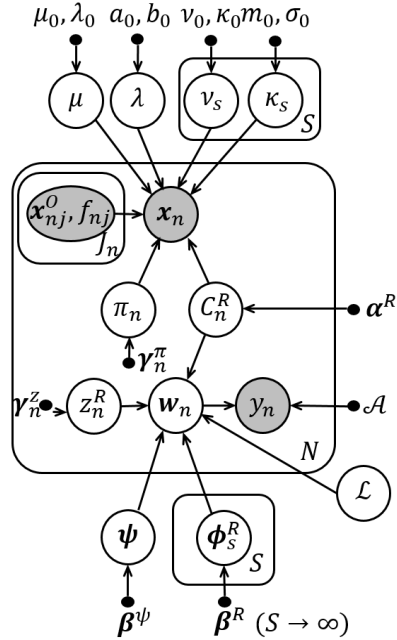

Figure 3. Graphical model of the proposed method ReSCAM.

Table 1. Distributions of ReSCAM

| | |
|---|---|
| $N(\cdot)$ | normal distribution |
| $vM(\cdot)$ | von Mises distribution |
| $U(\cdot)$ | uniform distribution |
| $Gam(\cdot)$ | gamma distribution |
| $logN(\cdot)$ | log-normal distribution |
| $Mult(\cdot)$ | multinomial distribution |
| $CRP(\cdot)$ | Chinese restaurant process |
| $Dir(\cdot)$ | Dirichlet distribution |

Table 2. Variables of ReSCAM

| | |
|---|---|
| $C_n^R$ | Index of relative spatial concepts |
| $\pi_n$ | Index of reference objects |
| $x_n$ | Absolute location of the trainer |
| $x'_{nj}$ | Location relative to a candidate reference object |
| $x_{nj}^O$ | Location of the candidate reference objects |
| $f_{nj}^O$ | Direction of the vector from candidate reference object to the robot |
| $\mu$ | Mean of distance |
| $\lambda$ | Precision of distance |
| $\nu_s$ | Mean angle of relative spatial concepts |
| $\kappa_s$ | Concentration of relative spatial concepts |
| $w_n$ | word sequences |
| $\phi_s^R$ | Parameter of word distribution of relative spatial concepts |
| $\psi$ | Parameter of word distribution of concept-independent words |
| $z_n^R$ | Index of location words |
| $y_n$ | Utterance |
| $\mathcal{A}$ | Acoustic model |
| $\mathcal{L}$ | Language model |
| $\mu_0, \lambda_0, a_0, b_0$ $\nu_0, \kappa_0, m_0, \sigma_0$ $e, \alpha^R, \boldsymbol{\beta}^R, \boldsymbol{\beta}^\psi$ $\gamma_n^\pi, \gamma_n^z$ | Hyperparameters |
| $N$ | Number of scenes |
| $S$ | Number of relative spatial concepts |
| $J_n$ | Number of candidate reference objects |

## 3.3 Iterative estimation procedure

This section describes the estimation procedure. A schematic diagram of the flow of iterative estimation is shown in Figure 4(**A**). We update the parameters in turn by iterating the four steps. A pseudo-code for the procedure is given in Algorithm 1.

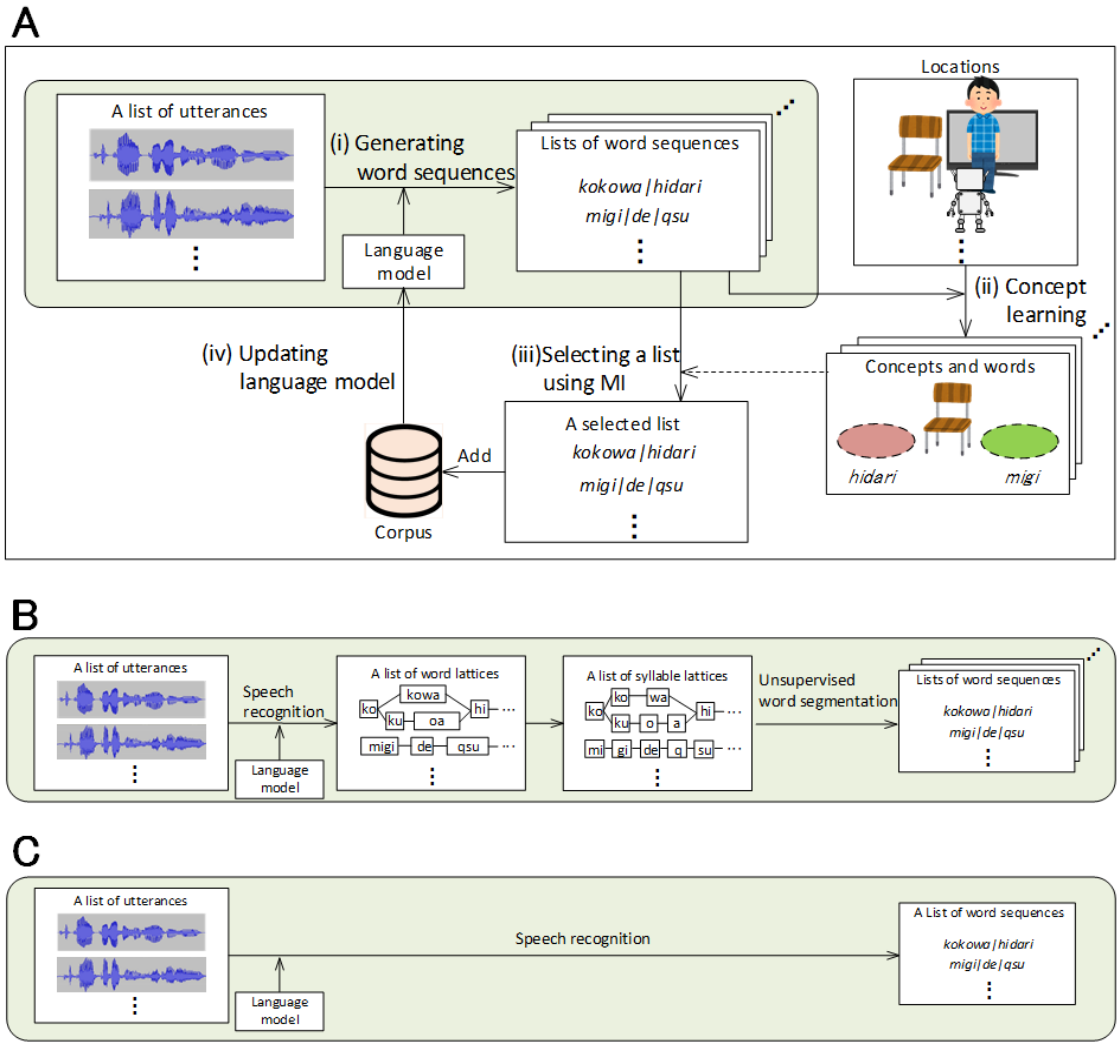

Figure 4. (**A**) Flow of the iterative estimation procedure. This iterative procedure consists of four steps. (i) A robot generates word sequences from utterances. (ii) Using the word sequences and locations, the robot estimates relative spatial concepts and location words. (iii) The robot selects the best list of word sequences using mutual information. (iv)The robot updates a language model using the selected list. (**B**) In iterations prior to the final iteration, word sequences are generated by speech recognition and unsupervised word segmentation. (**C**) In the final iteration, word sequences are generated only by speech recognition.

| Algorithm 1. Algorithm for learning of ReSCAM |
|---|
| Observe data in N scenes |
| Setting of hyperparameters |
| Initialize parameters |
| **for** iter = 1 to $G$ **do** |
|    (i) Generating word sequences               // Section 3.3.1 |
|    (ii) Concept learning                          // Section 3.3.2 |
|    (iii) Selecting a list using mutual information // Section 3.3.3 |
|    (iv) Updating of language model          // Section 3.3.4 |
| **end for** |
| |
| (i) Generating word sequences (final iteration) |
| (ii) Concept learning |

### 3.3.1 Generating word sequences

In iterations prior to the final iteration (Figure 4 (**B**)), utterances are converted to word lattices by speech recognition using language model $\mathcal{L}$, where the initial language model has only Japanese syllables. In order to use an unsupervised word segmentation method [20], the word lattices are then converted to syllable lattices by dividing the scores equally. After that, in order to obtain some candidates of word segmentation results, several lists of word sequences are generated from the converted syllable lattices using the unsupervised word segmentation method. In the final iteration (Figure 4 (**C**)), utterances were converted to word sequences using only speech recognition.

### 3.3.2 Concept learning

For lists of word sequences estimated as described in Section 3.3.1, parameters related to the concepts are learned using a list of word sequences $\boldsymbol{w} = \{\boldsymbol{w}_n\}_{n=1}^{N}$ and language model $\mathcal{L}$. Parameters of the posterior distributions $\boldsymbol{\nu}, \boldsymbol{\kappa}, \mu, \lambda, \boldsymbol{C}^R, \boldsymbol{\pi}, \boldsymbol{z}^R, \boldsymbol{\phi}^R,$ and $\boldsymbol{\psi}$ in equation (15) are estimated using the Metropolis–Hastings method, which is one of the Markov chain Monte Carlo (MCMC) sampling methods:

$$p(\boldsymbol{\nu}, \boldsymbol{\kappa}, \mu, \lambda, \boldsymbol{C}^R, \boldsymbol{\pi}, \boldsymbol{z}^R, \boldsymbol{\phi}^R, \boldsymbol{\psi} | \boldsymbol{x}', \boldsymbol{w}, \mathcal{L}, \boldsymbol{H}) \quad (15)$$

where $\boldsymbol{\nu} = \{\nu_1, \ldots, \nu_S\}, \boldsymbol{\kappa} = \{\kappa_1, \ldots, \kappa_S\}, \boldsymbol{C}^R = \{C_1^R, \ldots, C_N^R\}, \boldsymbol{\pi} = \{\pi_1, \ldots, \pi_N\}, \boldsymbol{z}^R = \{z_1^R, \ldots, z_N^R\}, \boldsymbol{x}' = \{\boldsymbol{x}_1', \ldots, \boldsymbol{x}_N'\}, \boldsymbol{w} = \{\boldsymbol{w}_1, \ldots, \boldsymbol{w}_N\},$ and $\boldsymbol{H} = \{\mu_0, \lambda_0, a_0, b_0, \nu_0, \kappa_0, m_0, \sigma_0, \alpha^R, \boldsymbol{\beta}^R, \boldsymbol{\beta}^\psi, \gamma_n^\pi, \gamma_n^z\}$. Each parameter is sampled in turn. The result of this step is a mean of samples obtained after burn-in.

### 3.3.3 Selecting a list based on mutual information

In the generating word sequences, there is a problem that the same word spoken in some utterances might be converted into several types of phoneme sequences by phoneme recognition errors. In SpCoA++ [6], a maximum mutual information criterion is used to solve a similar problem. In this method as well, we evaluate the adequacy of word sequences on the basis of a maximum mutual information criterion. Lists of word sequences are evaluated by mutual information using the parameters estimated from the lists. The mutual information between indices of relative spatial concepts and location words, denoted by $I^R$, is calculated as follows:

$$I^R = \sum_n \sum_s p(w_{nz_n^R}, C_n^R = s|\Theta) \log \frac{p(w_{nz_n^R}, C_n^R = s|\Theta)}{p(w_{nz_n^R}|\Theta)p(C_n^R = s|\Theta)}$$
$$= \sum_n \sum_s \phi_{sn}^R N_s \log \frac{\phi_{sn}^R}{\sum_{s'} \phi_{s'n}^R N_{s'}} \qquad (16)$$

where $N_s$ denotes the percentage of the data assigned to a relative spatial concept $s$, $\phi_{sn}^R$ denotes the value related to location word $w_{nz_n^R}$ in the word distribution parameter, and $\Theta$ denotes other parameters except $\boldsymbol{w}_n$, $z_n^R$, and $C_n^R$. A list of word sequences that maximizes the mutual information $I^R$ is selected to be added to the corpus.

### 3.3.4 Updating of language model

A class n-gram language model for speech recognition is calculated from the corpus mentioned above. The location word $w_{nz_n^R}$ in each scene is categorized into one class, and other words are categorized into each dependent class.

## 4  Relative spatial concept acquisition method using clues to reference objects (ReSCAM+O):

In the task setting explained in Section 3.1, no linguistic clues for use in estimating reference objects are given. This is a very difficult condition for the learner. On the other hand, in real interactions between humans, a speaker may include a word representing a reference object in an utterance or indicate it by using various types of information, such as a gesture, the direction of his/her gaze, and/or the context of the interaction. We considered the possibility that our robot could more flexibly learn relative spatial concepts by extending ReSCAM for such conditions. As an example of an extended ReSCAM, this section describes a model, called ReSCAM+O.

### 4.1  Task setting

We extended the work done in the previous task to consider more real interaction. Figure 5 shows an example of an interaction scene considered in the new task. In this task, the utterances of the trainer and object recognition by the robot were different from those in the previous task. In this task, the trainer always includes both words representing a relative concept and a reference object in his/her utterance. In addition, we assume that the robot has the capability for object recognition and can recognize each object as an object category. Under this condition, the robot simultaneously learns relative spatial concepts and words representing the relative spatial concepts and objects.

A word representing an object is a linguistic clue for estimating the reference object and makes relative concept learning efficient. However, the robot cannot directly know which object is the reference object because the robot does not know the relationship between object categories and words. Therefore, the robot needs to learn this relationship to use the linguistic clue. On the other hand, knowledge of relative concepts makes the learning of words representing objects more efficient because the robot can narrow down the candidate reference objects by understanding relative concepts included in utterances.

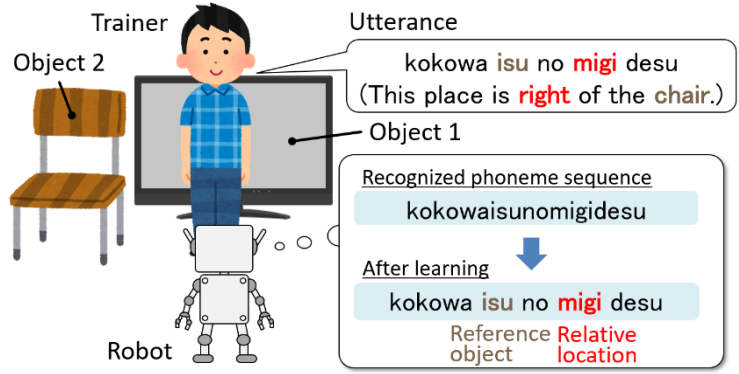

Figure 5. An interaction scene showing simultaneous lexical acquisition of relative spatial concepts and reference objects.

### 4.2 Probabilistic generative model

The graphical model is shown in Figure 6. The list of new variables is shown in Table 3. The new method differs from ReSCAM in the following ways. The object recognition result of reference objects in a scene $n$, denoted by $\boldsymbol{O}_{n\pi_n}$, is added to the observed parameters. It is represented as a one-hot vector. It is generated as follows:

$$\boldsymbol{O}_{n\pi_n} \sim \text{Mult}(\boldsymbol{\omega}_{C_n^O}) \qquad (17)$$

where $\boldsymbol{\omega} = \{\boldsymbol{\omega}_1, \ldots, \boldsymbol{\omega}_K\}$. To simplify the task, this object recognition result is assumed to be obtained by a virtual object recognition module without any recognition errors. The parameter $C_n^O, \boldsymbol{v}^O$ are generated as follows:

$$C_n^O \sim \text{Mult}(\boldsymbol{v}^O). \qquad (18)$$

$$\boldsymbol{v}^O \sim \text{Dir}(\boldsymbol{\alpha}^O). \qquad (19)$$

A word sequence $\boldsymbol{w}_n$ is generated as follows:
$$\boldsymbol{w}_n \sim p(\boldsymbol{w}_n | \boldsymbol{\phi}^R, \boldsymbol{\phi}^O, \boldsymbol{\psi}, C_n^R, C_n^O, z_n^R, z_n^O, \mathcal{L})$$

$$\stackrel{UR}{\approx} p(\boldsymbol{w}_n | \mathcal{L}) \prod_i \frac{p(w_{ni} | \boldsymbol{\phi}^R, \boldsymbol{\phi}^O, \boldsymbol{\psi}, C_n^R, C_n^O, z_n^R, z_n^O)}{p(w_{ni})}, \qquad (20)$$

where $\boldsymbol{\phi}^O = \{\boldsymbol{\phi}_1^O, \ldots, \boldsymbol{\phi}_K^O\}$. $p(w_{ni} | \boldsymbol{\phi}^R, \boldsymbol{\phi}^O, \boldsymbol{\psi}, C_n^R, C_n^O, z_n^R, z_n^O)$ denotes the prior probability of $w_{ni}$ and is calculated as follows:

$$p(w_{ni} | \boldsymbol{\phi}^R, \boldsymbol{\phi}^O, \boldsymbol{\psi}, C_n^R, C_n^O, z_n^R, z_n^O) = \begin{cases} \text{Mult}\left(w_{ni} | \boldsymbol{\phi}_{C_n^R}^R\right) & (i = z_n^R) \\ \text{Mult}\left(w_{ni} | \boldsymbol{\phi}_{C_n^O}^O\right) & (i = z_n^O) \\ \text{Mult}(w_{ni} | \boldsymbol{\psi}) & (\text{otherwise}) \end{cases} \qquad (21)$$

We define an *object word* as a word representing a reference object. The parameters of word distributions $\boldsymbol{\phi}_k^O$ and indices of object words $z_n^O$ are generated as follows:

$$\boldsymbol{\phi}_k^O \sim \text{Dir}(\boldsymbol{\beta}^O) \quad (22)$$

$$z_n^O \sim \text{Mult}(\boldsymbol{\gamma}^z) \quad (23)$$

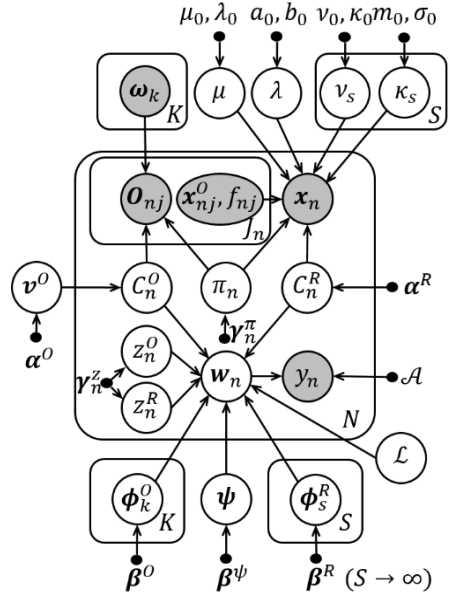

Figure 6. Graphical model of the proposed method ReSCAM+O

Table 3. New variables of ReSCAM+O

| | |
|---|---|
| $C_n^O$ | Index of object category |
| $\boldsymbol{\omega}_k$ | Parameter of distribution of object recognition result |
| $\boldsymbol{v}^O$ | Parameter of prior of index of object categories |
| $\boldsymbol{O}_{nj}$ | Object recognition result of each candidate of reference object |
| $\boldsymbol{\phi}_k^O$ | Parameter of word distribution of object category $k$ |
| $z_n^O$ | Index of object words |
| $\boldsymbol{\alpha}^O, \boldsymbol{\beta}^O$ | Hyperparameters |
| $K$ | Number of object categories |

### *4.3 Iterative estimation procedure*

The new iterative estimation procedure, shown in Figure 4(**A**) as with that used in ReSCAM, differs from that used in ReSCAM in the following ways. In step (iii), mutual information between indices of object categories and object words, $I^O$, is calculated as well as the mutual information $I^R$. Given that large values for the mutual information $I^R$ and $I^O$ mean that the relationship between a concept and a word is close to one-to-one, both values should be large. Therefore, a list of word sequences that maximizes $(I^O + I^R)$ is selected to be registered to the corpus. In step (iv), updating of the language model, the object word $w_{nz_n^O}$ in each scene is categorized into one class, as well as the location words.

## 5       Experiment I: ReSCAM

This experiment shows that relative spatial concepts can be learned using ReSCAM under the condition that clues to the reference objects are not provided. The task of this experiment is described in Section 3.1, and shown in Figure 2.

### *5.1     Conditions*

We taught a robot four relative spatial concepts shown in Table 4. We used a coordinate system, in which locations of candidate reference objects were set as the origin, and a direction from a reference object to the robot was set as a positive direction of the $y$ axis. A distance from the origin was set as a value sampled from a normal distribution. An angle between the $y$ axis and the line passing through the origin and the relative location was set as a value sampled from von Mises distribution. The parameters of distributions are shown in Table 4. Relative locations used as the training data are shown in Figure 7. We used utterances spoken in Japanese by one Japanese male speaker. The word sequences of the utterances were obtained using the parameters listed in Table 5. The 19 utterance patterns shown in Table 6 were used to simulate natural utterances. Examples of the utterances are listed in Table 7. We used Julius 4.5 and the Julius dictation-kit v4.4 [30][31] for speech recognition, and we used the acoustic model included in the Julius dictation-kit v4.4. The language model was not trained using any datasets in advance, and it only contained Japanese syllables with a uniform probability. We used latticelm v0.4 for unsupervised word segmentation. We set parameters as shown in Table 8.

We conducted experiments using the five methods listed in Table 9 to demonstrate the effectiveness of using mutual information and class n-grams, and to examine the upper bound of the performance. The hyperparameter values were set as follows: $\nu_0 = 0.0, \kappa_0 = 1.0, m_0 = 3.0, \sigma_0 = 0.3, \mu_0 = 0.0, \lambda_0 = 1.0, a_0 = 1.0, b_0 = 1.0, \alpha^R = 1.0, \boldsymbol{\beta}^R = \boldsymbol{\beta}^\psi = (0.1, \dots, 0.1)^T, \boldsymbol{\gamma}_n^\pi \sim (1.0, \dots, 1.0)^T$, and $\boldsymbol{\gamma}_n^z \sim (1.0, \dots, 1.0)^T$. Fifty trials were performed per method, with the numbers of candidate reference objects set to 3, 5, 10, 15, and 20.

Table 4.   Parameters used to generate relative locations

| Phoneme sequence | English | Directional distribution | | Distance distribution | |
|---|---|---|---|---|---|
| | | Mean angle [deg] | Concentration | Mean[m] | Standard deviation [m] |
| 'mae' | front | 0 | 14 | 1.0 | 0.20 |
| 'ushiro' | back | 90 | 14 | | |
| 'migi' | right | 180 | 14 | | |
| 'hidari' | left | 270 | 14 | | |

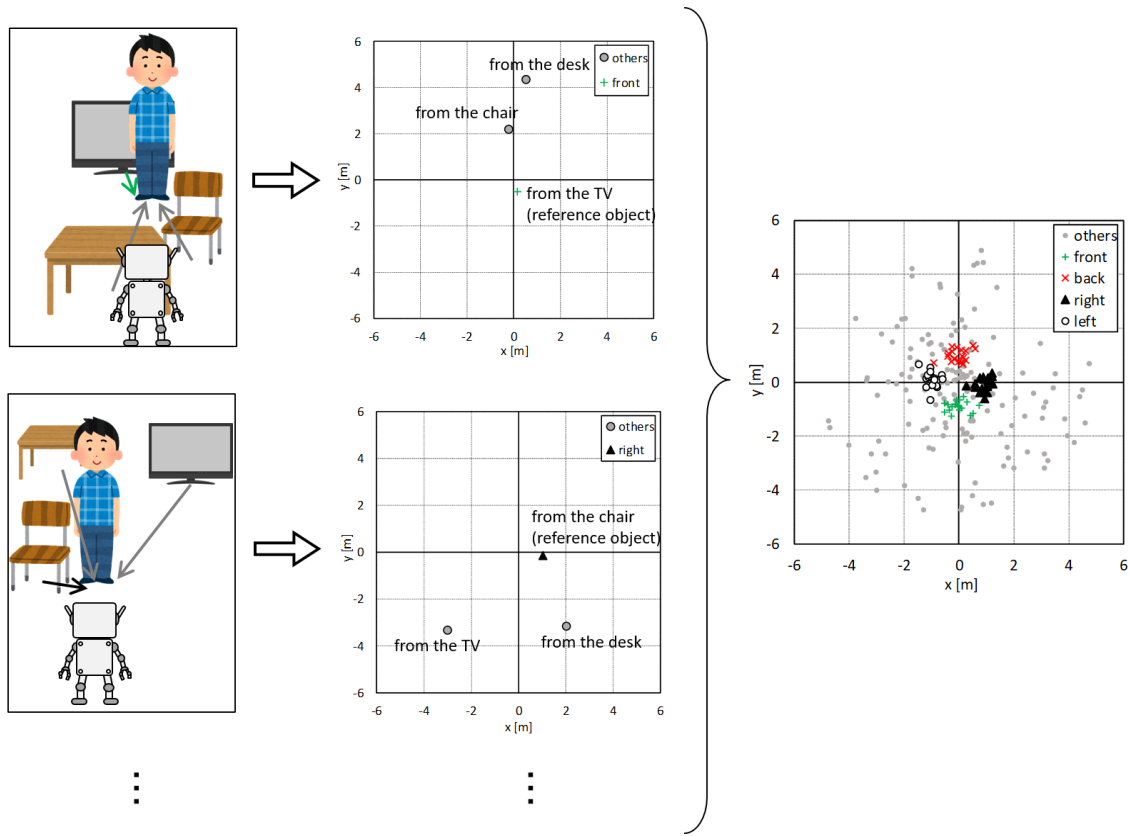

Figure 7. Relative locations used as training data. The origin represents the locations of the reference objects. Locations relative to all located objects in each scene are plotted, and the plotted points in all the scenes are combined in the figure on the right. The colored points represent the locations relative to the reference objects. The gray points represent the locations relative to the objects that are not reference objects. These locations were generated from a uniform distribution of distances from the reference objects from 0.1 to 5.0 m.

Table 5. Parameters used for generating utterances

| | |
|---|---|
| Number of relative spatial concepts | 4 |
| Number of utterances for each relative spatial concept | 19 |
| Number of speakers | 1 |
| Number of utterances | 76 |

Table 6. Utterance patterns used in the experiments. In '***', a phrase representing location is inserted. For example, 'migi', which means 'right', is inserted in the experiment I, and 'terebi no migi', which means 'right of the TV', is inserted in the experiment II.

| Utterance patterns | Translation to English |
|---|---|
| *** dane | It is ***. |
| *** dayo | It is ***. |
| *** desu | It is ***. |
| *** niirune | I am ***. |
| *** niiruyo | I am ***. |
| *** niimasu | I am ***. |
| *** nikimashita | I came ***. |
| kokowa *** | Here is ***. |
| kokononamaewa *** | The name of this place is ***. |
| konobashowa *** | This place is ***. |
| kokowa *** dane | Here is ***. |
| kokononamaewa *** dane | The name of this place is ***. |
| konobashowa *** dane | This place is ***. |
| kokowa *** dayo | Here is ***. |
| kokononamaewa *** dayo | The name of this place is ***. |
| konobashowa *** dayo | This place is ***. |
| kokowa *** desu | Here is ***. |
| kokononamaewa *** desu | The name of this place is ***. |
| konobashowa *** desu | This place is ***. |

Table 7. Examples of utterances. Bold words are location words.

| Word sequences | Translation to English |
|---|---|
| kokowa **mae** dane | Here is '**front**'. |
| konobashowa **migi** desu | This place is '**right**' |
| **hidari** niiruyo | I am in '**left**' |

Table 8. Parameters used for the experiment

| | |
|---|---|
| Number of iterations of iterative estimation | 20 |
| Burn-in period in concept learning | 50 |
| Number of iterations in concept learning | 100 |
| Length of class n-gram | 3 |
| Number of lists in selection by mutual information | 10 |

Table 9. A list of methods. In the CLM and Baseline methods, one list of word sequences is generated, and therefore a list of word sequences is not selected by mutual information. In the MI and Baseline methods, word 3-grams are generated instead of class 3-grams. In the Baseline method, semantic information does not influence the word sequences. The CLM+MI method is equal to the proposed method ReSCAM. The TRUEWORDS method is the concept learning where word sequences are given in order to examine the upper bound of the performance.

| Methods | Mutual information | Language model | Word sequences are given |
|---|---|---|---|
| Baseline | | word 3-gram | |
| MI | ✓ | word 3-gram | |
| CLM | | class 3-gram | |
| CLM+MI (ReSCAM) | ✓ | class 3-gram | |
| TRUEWORDS | | | ✓ |

## 5.2 Metrics

We evaluated the methods according to the following two metrics:

- Estimation accuracy of the relative spatial concepts, wherein we assessed the indices of the relative spatial concepts $C_n^R$ using the adjusted Rand index (ARI) [32], which is a measure of the degree of similarity between two clustering results.
- The phoneme accuracy rate (PAR) of the acquired words evaluates whether the learned phoneme sequences of the location words are properly segmented. PAR was calculated as the mean of the Levenshtein distance of the estimated location words and the correct words. As test data, we used 25 relative locations sampled from the true distribution of each relative spatial concept to calculate the PAR. Using each test data, the robot selects a location word that maximizes the posterior probability of location words[2]. We assumed that the location words were properly acquired if the robot could estimate correct phoneme sequences.

## 5.3 Results

We present the experimental results for the CLM+MI (ReSCAM) method for three candidate reference objects. Correct distributions and estimated distributions are shown in Figure 8(**A**), (**B**). The relative spatial concepts are learned with almost the same distributions as the correct distributions provided by ReSCAM. We consider that the differences between the estimated and correct distributions are caused by biases in the training data. After learning, the phoneme sequences listed in Table 10 were estimated. The estimated phoneme sequences are the exact words of relative spatial concepts. This shows that phoneme sequences can be learned using ReSCAM.

    We compare methods with increasing numbers of candidate reference objects. The ARI results are shown in Figure 9 (**A**), and the PAR results are shown in Figure 9

---
[2] We provide the details of the selection in Appendix A.

(**B**). For each number of candidate reference objects, we performed Tukey's tests with the results of the five methods. ARI and PAR for the CLM+MI method were the highest in the proposed methods for five or fewer candidate reference objects. The PARs for the CLM+MI and MI methods were significantly higher than the corresponding values for the Baseline method. This indicates that selecting word sequences using mutual information improves PAR. We discuss the reason for this result as follows. If selecting word sequences using mutual information is not performed, words with low appearance frequency may be generated due to phoneme recognition errors. In this case, wrong words may be associated with concepts because of low co-occurrence frequency between the phoneme sequences and the concepts. It is assumed that selecting word sequences using mutual information prevented the generation of low frequency words. In addition, for five or fewer candidate reference objects, the ARI for the CLM+MI method was significantly higher than those for the other proposed methods. Next, we discuss the effects of selecting word sequences using mutual information and class n-gram. By selecting word sequences using mutual information, the correct words were selected as words representing concepts, as mentioned above. This improves ARI because a concept uttered in each scene can be estimated using the word. In addition, using class n-gram, the word boundaries estimated while generating word sequences improve. For example, when a speech recognition result could be both of a correct segmentation 'kokowa / migi' and an incorrect segmentation 'kokowami / gi,' the former would have higher probability than the latter if the class of location words occurs after 'kokowa' frequently. This gap cannot be attributed to mutual information. This improves ARI as well as the selection of word sequences using mutual information.

When the number of candidate reference objects increased, ARI and PAR for all of the methods predictably decrease as a result of instability in the learning process caused by failure to estimate reference objects correctly. The accuracy rate of the index of the reference object $\pi_n$ is shown in Figure 9 (**C**). As with ARI and PAR, it decreased as the number of candidate reference objects increased. As shown in the results of TRUEWORDS, even if word sequences are provided in advance, the accuracy rates for estimating reference objects, ARI and PAR were low. Therefore, the performance is not improved using methods to improve word sequences. To learn in an environment with many candidate reference objects, other information for narrowing them down is needed.

Table 10. Examples of learning results for phoneme sequences.

| English | Correct | Estimated |
|---------|---------|-----------|
| front   | mae     | mae       |
| back    | ushiro  | ushiro    |
| right   | migi    | migi      |
| left    | hidari  | hidari    |

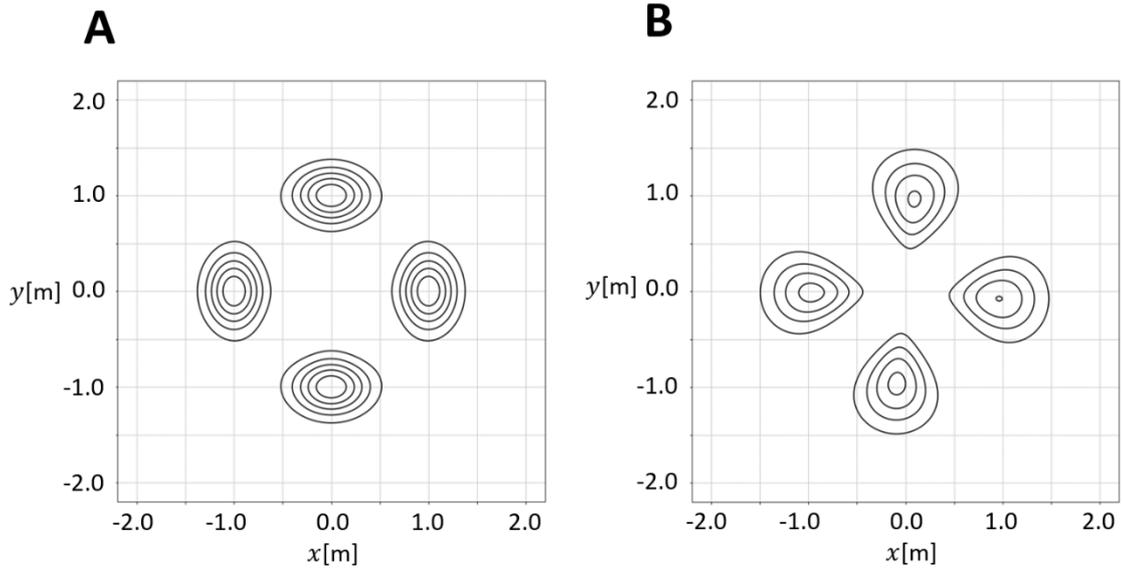

Figure 8. (**A**) Correct distributions. The training data were generated from the correct distributions. (**B**) Distributions estimated by the CLM+MI method for three candidate reference objects. The number of relative spatial concepts is correctly estimated as four.

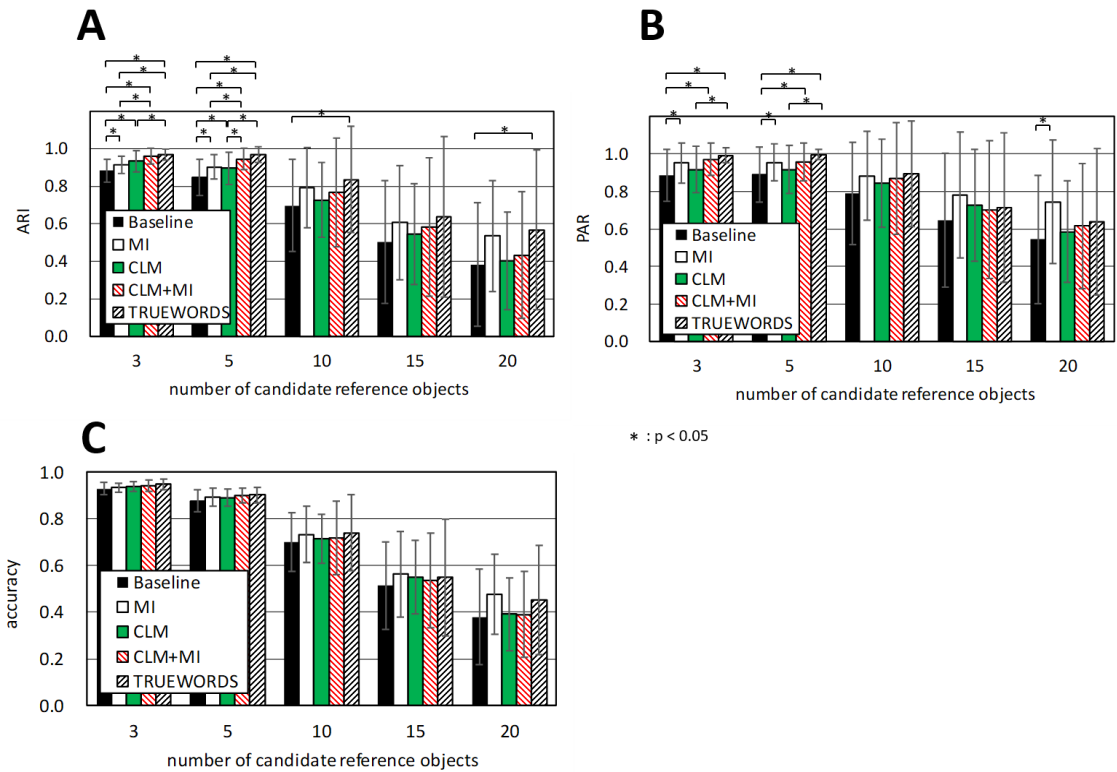

Figure 9. (**A**) ARI scores for estimating indices of relative spatial concepts. The thick bars represent the means, and the error bars represent the standard deviations. (**B**) PAR scores. (**C**) Accuracy rate of estimating reference objects.

## 6   Experiment II: ReSCAM+O

In this experiment, we demonstrate that giving clues to reference objects improves the estimation accuracy of relative spatial concepts when there are many candidate reference objects. This experiment is described in Section 4.1, and shown in Figure 5.

### 6.1   Conditions

The differences in conditions compared to those described in Section 5.1, were as follows. In addition to a location word, each utterance included an object word and either of two function words: 'no' or 'yori.' The examples are listed in Table 11. We used five object categories. The categories of objects that were not reference objects were randomly selected. The parameter $\boldsymbol{\omega}_k$ was set as a one-hot vector. To evaluate the effectiveness of using clues to reference objects, we compared the results of ReSCAM and ReSCAM+O. Hyperparameter values were set as follows: $\boldsymbol{\alpha}^O = (1.0, ..., 1.0)^T$ and $\boldsymbol{\beta}^O = (0.1, ..., 0.1)^T$.

Table 11.   Examples of utterances. Bold words are location words. Italic words are object words. The Japanese words 'no' and 'yori' are function words similar to 'of' in English.

| Word sequences | Translation to English |
| --- | --- |
| kokowa *terebi* no **mae** dane | Here is in **front** of the *TV*. |
| konobashowa *seNpuuki* no **migi** desu | This place is **right** of the *fan*. |
| *seNpuuki* yori **hidari** niiruyo | I am in **left** of the *fan*. |

### 6.2   Results

First, we show an example of the experimental results with ReSCAM+O for three candidate reference objects. The correct distributions and estimated distributions are shown in Figure 10(**A**), (**B**). The relative spatial concepts that were obtained by ReSCAM+O have almost the same distributions as the correct distributions. In addition, the estimated phoneme sequences are shown in Table 12 as with Section 5.3. The exact words of the relative spatial concepts were estimated. The phoneme sequences of the object words were estimated with high accuracy. This shows that phoneme sequences can be learned almost exactly using ReSCAM+O. Next, we compared the results for the three methods. For ARI and PAR for each number of candidate reference objects, we performed Student's t-tests with the results. The ARIs are shown in Figure 11 (**A**). For the three candidate reference objects, there is no difference in ARIs for ReSCAM and ReSCAM+O because relative spatial concepts can be acquired without clues when there are few candidate reference objects, as with the result in Section 5.3. For five or more candidate reference objects, the ARIs for ReSCAM+O were significantly higher than those for ReSCAM.

The PARs are shown in Figure 11 (**B**). As with the ARIs, there was no difference in the PARs for the three candidate reference objects. For 10 or more candidate reference objects, the PARs for ReSCAM+O were significantly higher than those for ReSCAM. Using ReSCAM, because relative spatial concepts cannot be learned, correct words related to locations cannot be selected. Therefore, the PARs for ReSCAM are low. On the other hand, ReSCAM+O can learn relative spatial concepts more accurately than ReSCAM. Therefore, correct words related to locations can be selected, and the PARs for ReSCAM+O are higher than those for ReSCAM.

The accuracy of estimating the reference objects is shown in Figure 11 (**C**). Using ReSCAM, as in Experiment 1, the accuracy of estimating reference objects decreases when the number of candidate reference objects increase. On the other hand, the decrease is small when using ReSCAM+O. We assume that these results show that relative spatial concepts can be learned using ReSCAM+O more accurately by improving the accuracy of estimating reference objects by extracting the linguistic clues of reference objects from utterances.

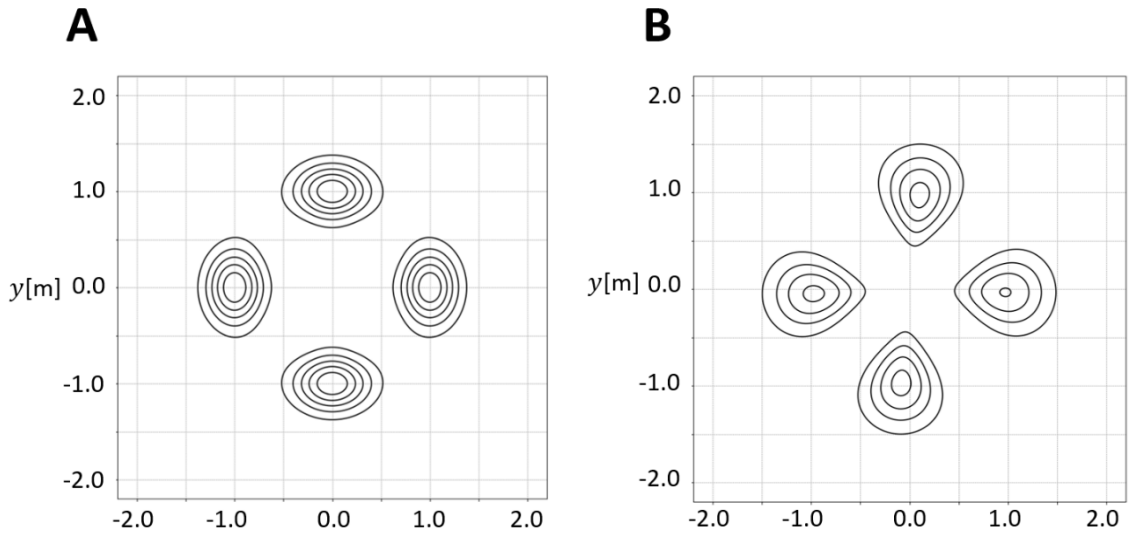

Figure 10. (**A**) Correct distributions. The training data were generated from the correct distributions. The number of relative spatial concepts was correctly estimated as four. (**B**) Distributions estimated by ReSCAM+O for three candidate reference objects.

Table 12. Learning results for phoneme sequences.

|  | English | Correct | Estimated | Accuracy of phoneme sequences |
|---|---|---|---|---|
| Relative spatial concepts | front | mae | mae | 1.00 |
|  | back | ushiro | ushiro | 1.00 |
|  | right | migi | migi | 1.00 |
|  | left | hidari | hidari | 1.00 |
| Objects | TV | terebi | terebi | 1.00 |
|  | fan | seNpuuki | seNpuki | 0.88 |
|  | screen | sukuriiN | sukuri | 0.75 |
|  | desk | tsukue | tsukue | 1.00 |
|  | PC | pasokoN | pasoko | 0.86 |

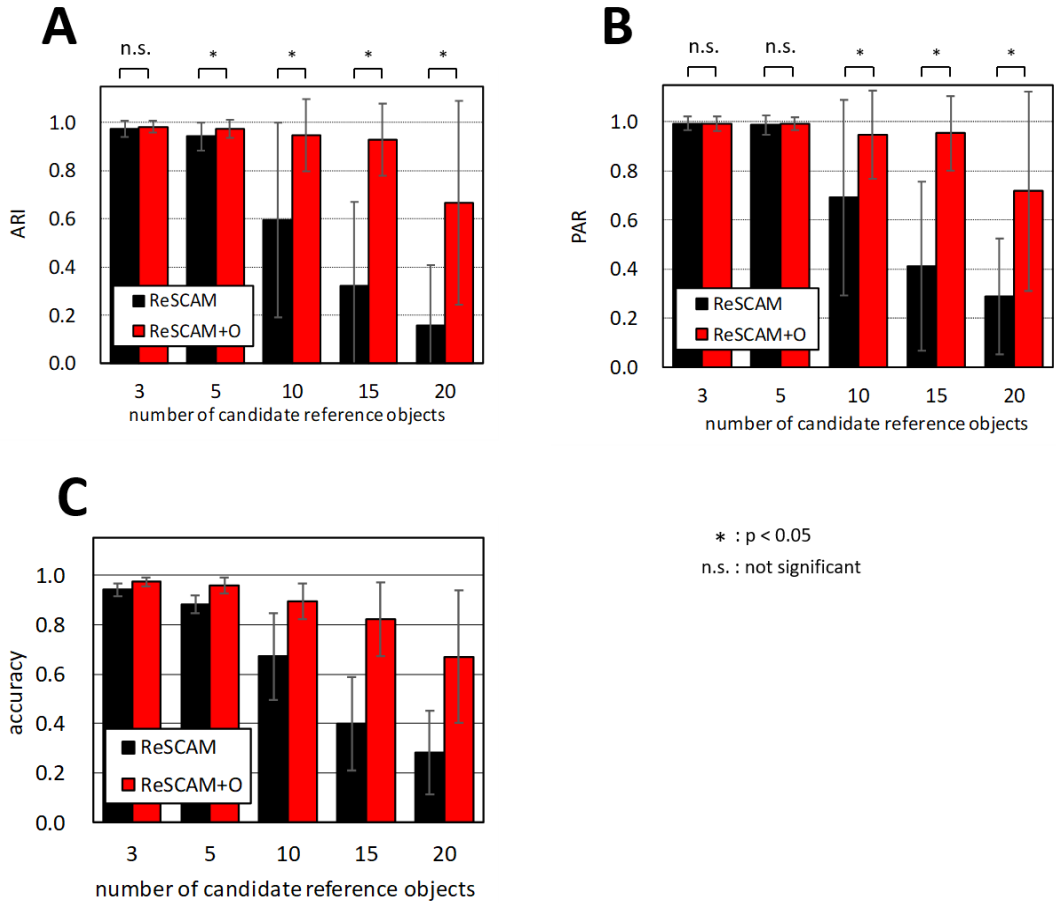

Figure 11. (**A**) ARI scores for estimating indices of relative spatial concepts. (**B**) PAR scores. (**C**) Accuracy rate of estimating reference objects.

## 7   Conclusions

In this paper, we proposed the unsupervised machine learning methods ReSCAM and ReSCAM+O for lexical acquisition of relative spatial concepts from spoken user utterances. The experimental results demonstrate that the proposed method ReSCAM is capable of estimating relative spatial concepts with high accuracy and segmenting the words related to the concepts without any clues to reference objects. The performance of unsupervised word segmentation from utterances was improved using class n-grams and mutual information. In addition, the experimental results show that the selecting word sequences using mutual information and using a class n-gram improve the estimation accuracy of relative spatial concepts. Furthermore, ReSCAM+O, which is extended from ReSCAM, is capable of improving the estimation accuracy of relative spatial concepts using clues to reference objects. This shows that we can develop methods using other observation as clues to reference objects (e.g., gestures and context).

In our experiments, object recognition results were not obtained in an actual environment. We also plan to develop a method for learning concepts using recognition results obtained from real objects in order for our method to be practical.

Additionally, we plan to address other relative concepts, such as large and small brightness and colors. A robot would could understand many spoken user utterances by combining learned concepts if other relative concepts can be learned.

**Funding**

This work was supported by an MEXT Grant-in-Aid for Scientific Research on Innovative Areas under Grant JP16H06569.

**References**

[1] Chomsky N. Rules and Representations. Oxford: Basil Blackwell; 1980.
[2] Quine WV. Word and Object. Cambridge, MA.: MIT Press, 1960.
[3] Tomasello M. The Cultural Origins of Human Cognition. Harvard University Press, 1999.
[4] Taniguchi T, Nagai T, Nakamura T, et al. Symbol Emergence in Robotics: A Survey. Adv Robot. 2016;30(11-12):706-728.
[5] Taniguchi T, Mochihashi D, Nagai T, et al. Survey on frontiers of language androbotics. Adv Robot. 2019;33(15-16):700-730.
[6] Taniguchi A, Taniguchi T, Inamura T. Unsupervised spatial lexical acquisition by updating a language model with place clues. Robot Auton Syst. 2018;99:166–180.
[7] Taniguchi A, Hagiwara Y, Taniguchi T, et al. Online Spatial Concept and Lexical Acquisition with Simultaneous Localization and Mapping. 2017 IEEE/RSJ International Conference on Intelligent Robots and Systems; 2017 September 24-28; Vancouver, Canada.
[8] Taniguchi A, Hagiwara Y, Taniguchi T, et al. Improved and scalable online learning of spatial concepts and language models with mapping. Auton Robot. 2020;44(6):927-946.
[9] Taniguchi A, Hagiwara Y, Taniguchi T, et al. Spatial Concept-Based Navigation with Human Speech Instructions via Probabilistic Inference on Bayesian Generative Model. Adv Robot. 2020;34(19):1213-1228.
[10] Frank M, Goodman N, Tenenbaum JB. A Bayesian framework for cross-situational word-learning. Adv Neural Inf Process Syst. 2008;20:457–464.
[11] Heath S, Ball D, Wiles J. Lingodroids: Cross-Situational Learning for Episodic Elements. IEEE Trans Cogn Devel Syst. 2016;8(1):3–14.
[12] Hatori J, Kikuchi Y, Kobayashi S, et al. Interactively picking real-world objects with unconstrained spoken language instructions. Proceedings of International Conference on Robotics and Automation (ICRA2018). 2018 May 21-26; Brisbane, Australia. p. 1-6.
[13] Štěpánová K, Klein FB, Cangelosi A, et al. Mapping language to vision in a real-world robotic scenario. IEEE Trans Cogn Devel Syst 2018;10:784–794.

[14] Taniguchi A, Taniguchi T, Cangelosi A. Cross-situational learning with Bayesian generative models for multimodal category and word learning in robots. Front Neurorobotics. 2017;19:66.

[15] Heymann J, Walter O, Haeb-Umbach R, et al. Iterative Bayesian word segmentation for unsupervised vocabulary discovery from phoneme lattices. The 2014 IEEE International Conference on Acoustics, Speech and Signal Processing (ICASSP); 2014 May 4–9; Florence, Italy.

[16] Synnaeve G, Dautriche I, Börschinger B, et al. Unsupervised word segmentation in context. Proceedings of COLING 2014, the 25th International Conference on Computational Linguistics; 2014 August 23–29; Dublin, Ireland.

[17] Araki T, Nakamura T, Nagai T, et al. Online learning of concepts and words using multimodal LDA and hierarchical Pitman-Yor language model. 2012 IEEE/RSJ International Conference on Intelligent Robots and Systems; 2012 October 7-12; Vilamoura, Portugal.

[18] Nakamura T, Nagai T, Funakoshi K, et al. Mutual learning of an object concept and language model based on MLDA and NPYLM. In the IEEE International Conference on Intelligent Robots and Systems; 2014 September 14–18; Chicago, IL, USA.

[19] Mochihashi D, Yamada T, Ueda N. Bayesian unsupervised word segmentation with nested Pitman-Yor language modeling. ACL-IJCNLP 2009, Joint Conference of the 47th Annual Meeting of the Association for Computational Linguistics and 4th International Joint Conference on Natural Language Processing of the AFNLP; 2009 August 2-7; Singapore.

[20] Neubig G, Mimura M, Mori S, et al. Bayesian learning of a language model from continuous speech. IEICE T Inf Syst. 2012;E95.D(2):614–625.

[21] Sugiura K, Iwahashi N, Kashioka H, et al. Learning, generation and recognition of motions by reference-point- dependent probabilistic models. Adv Robotics. 2011;25:825–848.

[22] Gu Z, Taguchi R, Hattori K, et al. Learning of relative spatial concepts from ambiguous instructions. IFAC-PapersOnLine. 2016;49:19.

[23] Barrett DP, Bronikowski SA, Yu H, et al. Robot language learning, generation, and comprehension. 2015 [cited 2021 Feb 8];[10 p.]. Available from: https://arxiv.org/abs/1508.06161.

[24] Aly A, Taniguchi A, Taniguchi T. A generative framework for multimodal learning of spatial concepts and object categories: An unsupervised part-of-speech tagging and 3D visual perception based approach. 7th Joint IEEE International Conference on Development and Learning and on Epigenetic Robotics, ICDL-EpiRob 2017; 2018 April 5; Lisbon, Portugal.

[25] Spranger M. Grounded lexicon acquisition - Case studies in spatial language. 2013 IEEE Third Joint International Conference on Development and Learning and Epigenetic Robotics (ICDL). 2013 August 18-22; Osaka, Japan.

[26] Spranger M. Incremental grounded language learning in robot-robot interactions: Examples from spatial language. 2015 Joint IEEE International Conference on

Development and Learning and Epigenetic Robotics (ICDL-EpiRob); 2015 August 13–16; Providence, RI, USA.
[27] Qin X, Padraig C, Michael S. Online Trans-Dimensional von Mises-Fisher Mixture Models for User Profiles. J Mach Learn Res. 2016;17(200):1-51.
[28] Pitman J. Exchangeable and partially exchangeable random partitions. Probab Theory Relat Fields. 1995;102:145–158.
[29] Gildea D, Hofmann T. Topic-based language models using EM. Proceedings of European Conference on Speech Communication and Technology (EUROSPEECH); 1999 September 5-9; Budapest, Hungary. p. 2167-2170.
[30] Kawahara T, Kobayashi T, Takeda K, et al. Sharable software repository for Japanese large vocabulary continuous speech recognition. Proceedings of the ICSLP 1998, the 5th International Conference on Spoken Language Processing; 1998 November 30–December 4; Sydney, Australia.
[31] Lee A, Kawahara T, Shikano K. Julius --- An open source real-time large vocabulary recognition engine. Proceedings of European Conference on Speech Communication and Technology (EUROSPEECH); 2001 September 3-7; Aalborg, Denmark. p. 1691–1694.
[32] Hubert L, Arabie P. Comparing partitions. J Classif. 1985;2:193–218.

**Appendix**

### A    *Estimation of location words and object words from new sensory input*

After learning of ReSCAM or ReSCAM+O, the robot is able to indicate a location to the user by uttering location words. First, the robot arbitrarily selects a reference object. Next, the robot selects the location word $w_n^R$ representing the location relative to the reference object, $x_n'$, as follows:

$$\begin{aligned}
w_n^R &= \arg\max_w p\left(w \middle| x_n', \Theta\right) \\
&= \arg\max_w \sum_s p(w|\phi_s^R) p(C_n^R = s | \alpha^R) p\left(x_n' \middle| \mu, \lambda, \nu, \kappa, \pi_n, C_n^R = s\right)
\end{aligned} \quad (24)$$

In this equation, the word that maximizes the posterior probability of location words is selected.

After learning of ReSCAM+O, the robot also selects an object word of the reference object $w_n^O$ from object recognition results $O_n$ as follows:

$$\begin{aligned}
w_n^O &= \arg\max_w p(w|O_n, \Theta) \\
&= \arg\max_w \sum_k p(w|\phi_k^O) p(C_n^O = k | \nu^O) p\left(O_{n\pi_n} \middle| \omega, C_n^O = k\right)
\end{aligned} \quad (25)$$

## B  Verification of hyperparameters

We set the hyperparameters to tolerate the change in the training data to some extent. For example, we show that learning can be performed when distances between the reference objects and the locations of the trainer change to some extent in the setting of the Experiment I. In this experiment, the mean parameter, which is used in normal distribution to generate the distances, were set from 1.0 to 10.0 in every 1.0 step instead of setting to 1.0 as in the case of the Experiment I. We also performed experiments where the burn-in period was set to 500, and the number of iterations was set to 1000 in the Metropolis–Hastings method used in concept learning. Ten trials are performed per condition. The means of ARI and PAR are shown in Fig.B1. First, we discuss the result with 100 iterations as in the case of the Experiment I. When the distance was smaller than six, the performances were almost the same as that when the distance was one. However, the performance deteriorated when the distance was larger than six. This shows that the hyperparameters relevant to distances are set so that it can tolerate the change of training data to some extent. As expected, the values of the hyperparameters must be adopted to the training data when the condition of the training data changes significantly. Next, we discuss the result with 1000 iterations. When the distances are large, performances are better than that in the case of 100 iterations. We believe that this is because the sampling period to converge is extended by the difference between the prior distribution determined by the hyperparameters and distances used to generate the training data.

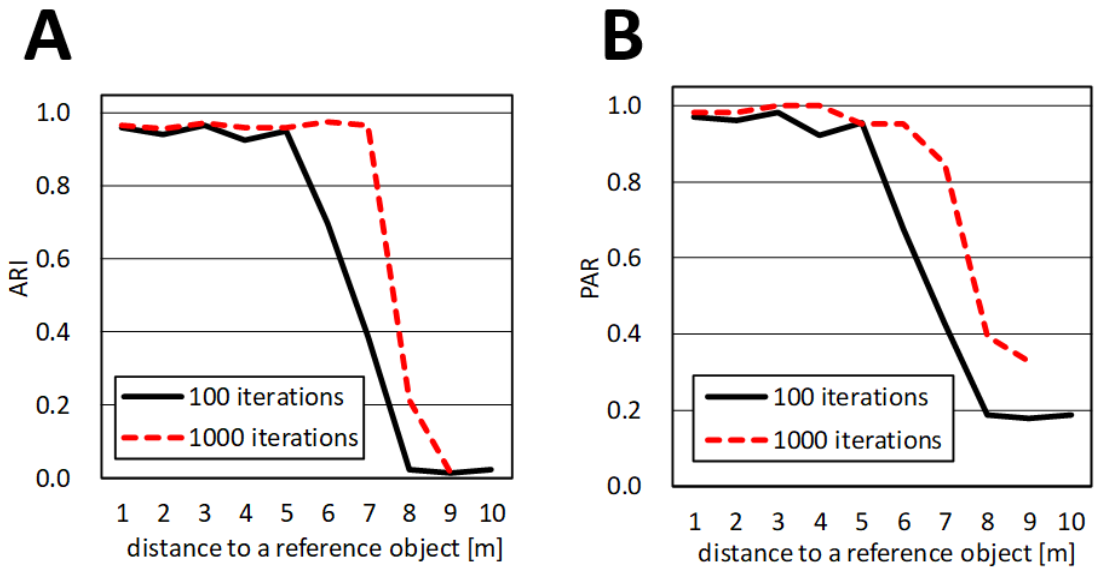

Figure B1. Scores of each distance to a reference object. We performed experiments in two conditions; 100 and 1000 iterations of Metropolis–Hastings method in concept learning. (**A**) ARI scores. (**B**) PAR scores.